\documentclass[11pt,a4paper,english,conference]{IEEEtran}
\usepackage[utf8]{inputenc}
\usepackage{mathtools}
\usepackage{amsmath}
\usepackage{amssymb}
\usepackage{graphicx}

\makeatletter

\pdfpageheight\paperheight
\pdfpagewidth\paperwidth

\@ifundefined{date}{}{\date{}}

\usepackage{times}
\usepackage{epsfig}
\usepackage{graphics}

\makeatother

\usepackage{babel}
\begin{document}

\title{Facial Expression Recognition Using Sparse Gaussian Conditional Random
Field}

\author{\IEEEauthorblockN{Mohammadamin Abbasnejad} \IEEEauthorblockA{School of Electrical and Computer Engineering\\
 Shiraz University\\
 Shiraz, Iran\\
 Email: amin.abbasnejad@gmail.com} \and \IEEEauthorblockN{Mohammad Ali Masnadi-Shirazi} \IEEEauthorblockA{School of Electrical and Computer Engineering\\
 Shiraz University\\
 Shiraz, Iran\\
 Email: mmasnadi@shirazu.ac.ir} }

\maketitle



\begin{abstract}
The analysis of expression and facial Action Units (AUs) detection
are very important tasks in fields of computer vision and Human Computer
Interaction (HCI) due to the wide range of applications in human life.
Many works has been done during the past few years which has their
own advantages and disadvantages. In this work we present a new model
based on Gaussian Conditional Random Field. We solve our objective
problem using ADMM and we show how well the proposed model works.
We train and test our work on two facial expression datasets, CK+
and RU-FACS. Experimental evaluation shows that our proposed approach
outperform state of the art expression recognition. \end{abstract}


\section{Introduction}

\label{introduction} Over the past few years the problem of temporal
classification and recognition has gained significant attention among
researchers in many fields such as speech recognition, human expression
classification, event detection and etc. Generally in temporal analysis
the goal is to find a mapping function from the stream of observation
sequences to the set of corresponding outputs. There are many algorithms
that tackle this problem: in \cite{rabiner1986introduction}, the
authors tackled speech recognition problem by using Hidden Markov
Model (HMM). HMM is a well known classifier that by modeling the joint
probability of inputs $\mathbf{x}$, and outputs $\mathbf{y}$ conditioned
on a set of latent state variable learns a temporal transition. Although
HMM is widely used and has been reported to work well in similar applications,
there are pitfalls in using it. One of the main problems is that HMM
predicts a joint probability distribution between inputs and outputs.
To define a joint probability distribution between observations and
labels, HMM needs to model distribution of $p(\mathbf{x})$ which
can include complex dependencies. Modeling these dependencies in the
input make it complex, however disregarding it reduces performance.

The solution to this problem is to instead of modeling joint probability
distribution, directly model the conditional distribution $p(\mathbf{x}|\mathbf{y})$.
This approach is exactly the conditional random field {\cite{lafferty2001conditional}}.
The conditional random field is a graphical model that maximizes the
conditional probability distribution between inputs and outputs. This
assumption makes CRF to have much more rich feature observation than
HMM. However CRF plays an important role in many computer vision applications
but in general, parameter learning and inference are very complex
and time consuming because of nonlinearity and non-convexity of CRF.
In \cite{zhu1998filters,roth2005fields} sampling algorithms for parameters
learning are used, but unfortunately, sampling algorithms are slow
to convergence.

Gaussian Markov random field is the ordinary MRF model where variables
are jointly Gaussian. Gaussian models are very popular in many fields
of studies because of the inference in Gaussian models can be easy.
Typically, the key success with Gaussian MRF as indicated in \cite{tappen2007learning,wytock2013sparse}
is the neighboring functions that are dependent on the input signal.
This dependency among input signals makes GMRF to be a conditional
model and can be called Gaussian Conditional Random Field (GCRF).

In this paper we introduce a model based on GCRF and show how the
parameters of this model can be efficiently learned. The model we
propose is based on ADMM model  {\cite{boyd2011distributed}}. The
rest of this paper is organized as follow: in section 2 we are going
to introduce an introduction about GCRF, in section 3 our proposed
model is presented and in section 4 and 5 our experimental results
and conclusion are introduced respectively.

\section{Gaussian Conditional Random Field}


In this section we are going to present an introduction about GCRF.
A GCRF model can be formalized as follow: 
\begin{equation}
p(\mathbf{y}|\mathbf{x};\Theta,\Lambda)=\frac{1}{Z(\mathbf{x})}\exp\{-\mathbf{y}^{T}\Lambda\mathbf{y}-2\mathbf{x}^{T}\Theta\mathbf{y}\}\label{eq:eq1}
\end{equation}
where in this equation~$\mathbf{x}\in\Re^{n}$is the sequence of
observation inputs and $\mathbf{y}\in\Re^{p}$ is the sequence of
corresponding outputs. In this equation $\Lambda$ is a parameter
which model conditional dependency among $\mathbf{y}$ and $\Theta$
maps the input to output. The partition function is given by: 
\begin{equation}
\frac{1}{Z(\mathbf{x})}=c|\Lambda|\exp\{-\mathbf{x}^{T}\Theta\Lambda^{-1}\mathbf{x}\}\label{eq:eq2}
\end{equation}
In this model the maximum likelihood is given by getting negative
log of Eq.\ref{eq:eq1} and Eq.\ref{eq:eq2} and is defined as follow:
\begin{equation}
f(\Lambda,\Theta)=-\log{|\Lambda|}+\text{tr}\{S_{yy}\Lambda+2S_{yx}\Theta+\Lambda^{-1}\Theta^{T}S_{xx}\Theta\}\label{eq:eq3}
\end{equation}
where in this equation, terms $S$ are the empirical covariances and
are: 
\begin{equation}
S_{yy}=\frac{1}{m}\mathbf{y}^{T}\mathbf{y},S_{yx}=\frac{1}{m}\mathbf{y}^{T}\mathbf{x},S_{xx}=\frac{1}{m}\mathbf{x}^{T}\mathbf{x}\label{eq:eq4}
\end{equation}

As indicated in \cite{wytock2013sparse} this equation is a convex
problem and solution to this problem can be done by getting gradient
with respect to parameters and using steepest descent method to find
the parameters which maximize Eq.\ref{eq:eq4}. Unfortunately using
steepest descent method is very computational and it needs many numbers
of iterations for convergence. In this paper we are going to present
a new method based on ADMM and show how fast we can learn parameters
from the model.


\section{Proposed Method}


As can be seen from the previous section, the main problem of GCRF
is complexity and time consumption. In this section we are going to
introduce our proposed method based on ADMM model after an introduction
about ADMM.

\subsection{Alternating Direction Method of Multipliers}

\label{sec:proposed_method} ADMM which first introduced by \cite{boyd2011distributed}
is a robust and fast optimization method. Consider the optimization
problem such as: 
\begin{eqnarray}
\min_{\mathbf{x},\mathbf{z}}\quad f(\mathbf{x})+g(\mathbf{z}),\nonumber \\
\text{s}.\text{t}.\quad\mathbf{A}\mathbf{x}+\mathbf{B}\mathbf{z}=c\label{eq:eq5}
\end{eqnarray}
where for variables $\mathbf{x},\mathbf{z}\in{R^{n}}$, two functions
$f$,$g$ are convex. The augmented Lagrangian of Eq. \ref{eq:eq5}
is defined: 
\begin{eqnarray}
L_{\rho}(\mathbf{x},\mathbf{z},q) & = & f(\mathbf{x})+g(\mathbf{z})+q^{T}(\mathbf{A}\mathbf{x}+\mathbf{B}\mathbf{z}-c)\nonumber \\
 &  & \quad+\frac{\rho}{2}\|\mathbf{A}\mathbf{x}+\mathbf{B}\mathbf{z}-c\|_{2}^{2}\label{eq:eq6}
\end{eqnarray}
where $\rho$ is a positive and tunable parameter. The $k$th iteration
of ADMM technique can be defined as follows: 
\begin{eqnarray}
\mathbf{x}^{k+1} & = & \arg\min_{x}L_{\rho}(\mathbf{x},\mathbf{z}^{k},\mathbf{q}^{k})\nonumber \\
\mathbf{z}^{k+1} & = & \arg\min_{z}L_{\rho}(\mathbf{x}^{k+1},\mathbf{z},\mathbf{q}^{k})\\
{q}^{k+1} & = & \mathbf{q}^{k}+\rho(\mathbf{A}\mathbf{x}^{k+1}+\mathbf{B}\mathbf{z}^{k+1}-c)\nonumber 
\end{eqnarray}
these three steps are iteratively done until convergence. From above
steps we can see that if we minimize over $\mathbf{x}$ and $\mathbf{z}$
the method reduces to the classic method of multipliers. Instead ADMM
by fixing the variable $\mathbf{z}$ and minimizing over $\mathbf{x}$
and vice versa, finding the optimum value for the optimization problem.
This assumption makes ADMM to be more robust and the number of iterations
for convergence decreases dramatically.

\subsection{Proposed Method}

\label{sec:proposed_method} As can be seen from last subsection ADMM
is a fast and accurate optimization method which we are going to use
it for our approach. From Eq.{\ref{eq:eq3}, Eq.{\ref{eq:eq5} we
can rewrite the GCRF as follows: 
\begin{eqnarray}
f(\Lambda,\Theta,\Phi) & = & -\log{|\Lambda|}+tr\{S_{22}\Lambda+2S_{12}\Theta+S_{11}\Phi\}\nonumber \\
\text{s}.\text{t}\quad\Phi & = & \Theta^{T}\Lambda^{-1}\Theta\label{eq:eq8}
\end{eqnarray}
We propose to handle Eq. \ref{eq:eq8} by using ADMM \cite{boyd2011distributed}.
The ADMM-style for the optimization problem takes the following form
for the Lagrangian:{\footnotesize{}
\begin{eqnarray*}
L(\Lambda,\Theta,\Phi,\mathbf{q)}\!\!\!\!\! & =\!\!\!\!\!\! & -\log{|\Lambda|}+\frac{\mu}{2}\|\Phi-\Theta^{T}\Lambda^{-1}\Theta\|_{2}^{2}\\
 &  & \,+\text{tr}(S_{22}\Lambda+2S_{12}\Theta+S_{11}\Phi)+\mathbf{q}^{T}(\Phi-\Theta^{T}\Lambda^{-1}\Theta)
\end{eqnarray*}
}where $\mathbf{q}$ is the Lagrangian vector and $\mu$ is the penalty
factor that controls the rate of convergence. The ADMM iteration steps
are forms as: {\footnotesize{}
\begin{eqnarray*}
\Lambda^{*} & = & -\Lambda^{-1}+S_{22}+(\Lambda^{-1}\Theta^{T}\Theta\Lambda^{-1})(\mathbf{q}^{T}+\mu(\Phi-\Theta^{T}\Lambda^{-1}\Theta))\\
\Theta^{*} & = & 2S_{12}-\Theta\Lambda^{-1}(2\mathbf{q}^{T}+\mu(\Phi-\Theta^{T}\Lambda^{-1}\Theta))\\
\Phi^{*} & = & S_{11}+q^{T}+\mu(\Phi-\Theta^{T}\Lambda^{-1}\Theta)
\end{eqnarray*}
}The Lagrangian vector update in each iteration as follows: 
\begin{equation}
\mathbf{q}^{i+1}\leftarrow\mathbf{q}^{i}+\mu(\Lambda^{*(i+1)}-\Theta^{*(i+1)}-\Phi^{*(i+1)})\label{eq:eq10}
\end{equation}
A simple and common \cite{boyd2011distributed} scheme for selecting
$\mu$ is following:

\begin{equation}
\mathbf{\mu}^{i+1}=\arg\min(\mu_{max},\beta\mu^{i})\label{eq:eq10}
\end{equation}
we found experimentally $\mu^{0}=10^{-2},\beta=1.1$ and $\mu_{max}=20$
to perform well.

\section{Feature extraction from video}

This section describe the feature extraction method at frame-level
and representing the extracted features in each segment.

\subsection{Facial feature extraction}

There are two general approaches for video feature extraction, shape-based
\cite{carlsson2001action,valstar2005facial} and appearance-based
\cite{zhu2009dynamic,sikka2012exploring} methods. Common to all appearance-based
methods, they have some limitations due to changes in camera view,
illumination and speed of action. On the other hand, geometrical approaches
by following the movement over some key parts (on body or face) try
to capture the temporal movement in a sequence of observations. In
this paper, we use the shape technique to represent each video frame
vector. We use facial feature points and 6D comprehensive motion data,
including position, orientation, acceleration and angular speed tracking
for body gesture to build the observation data. The facial points
are tracked using Constrained Local Models (CLM) \cite{asthana2013robust}.
After the facial components have been tracked, a similarity transformation
is applied to facial features with respect to the normal facial shape
to eliminate all variations including, scale, rotation and transition.
Figure \ref{fig:database} shows an example of facial landmark features
in several frames of RU-FACS \cite{dibekliouglu2012you} video database.

\section{Experiment setting and Databases}

This section describes our experiments on two publicly available dataset,
CK+ \cite{lucey2010extended} and RU-FACS \cite{bartlett2006automatic}
Dataset. 

\subsection{Datasets}

\textbf{CK+ Database} The CK+ Database is a facial expression database.
This database contains 593 facial expression sequences from 123 participants.
Each sequence starts from neutral face and ends at the peak frame.
Sequences vary in duration between 4 and 71 frames and the location
of 68 facial landmarks are provided along with database. Facial pose
is frontal with slightly head motion. All the facial feature points
were registered to a reference face by using similarity transformation.
Some examples from this database is shown in Figure \ref{fig:database}

\begin{figure}[t]
\begin{centering}
\label{database} \includegraphics[scale=0.7]{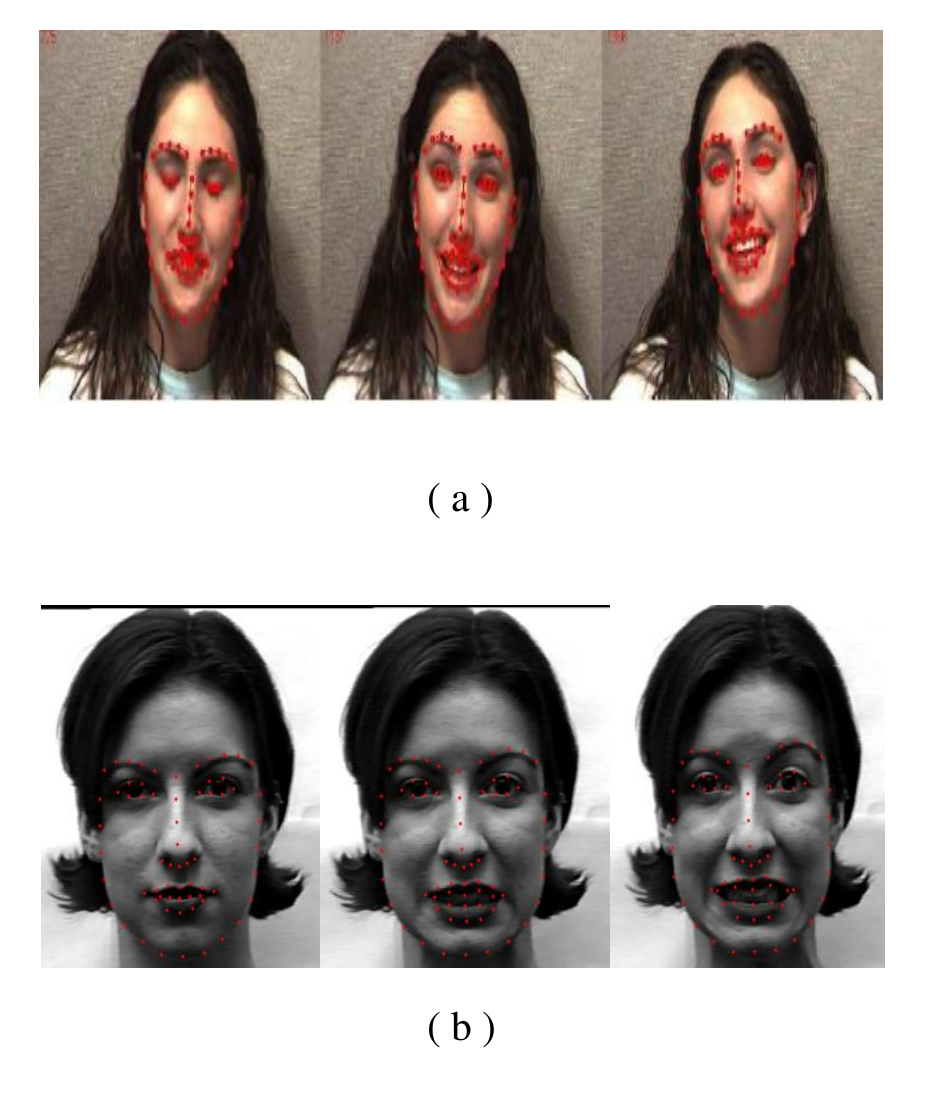} 
\par\end{centering}

\caption{a) Some examples for RU-FACS database, b) Some examples for CK+ databases}

\label{fig:database} 
\end{figure}

\begin{figure*}
\begin{centering}
\label{result1} \includegraphics[scale=0.8]{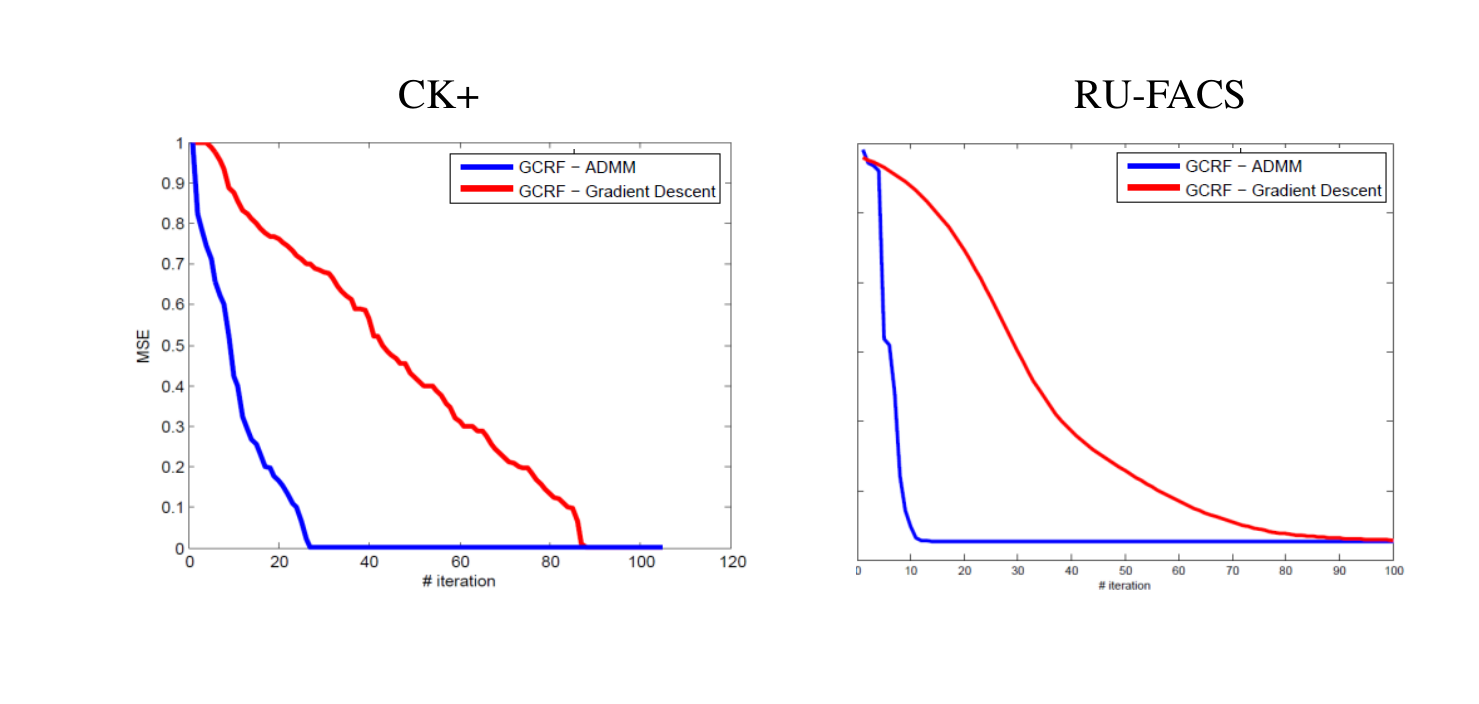} 
\par\end{centering}

\caption{Convergence with respect to number of iterations for each datset}

\label{fig:result1}
\end{figure*}

is more an expression dataset and is more challenging than the CK+
dataset, and it consists of facial behavior recorded during interviews.
The interviews are about two minutes. Participants show moderate pose
variation and speech-related mouth movements. Compared with the CK+
datasets, RU-FACS is more natural in timing, much longer, and the
AUs are at lower intensity. For technical reasons, we selected from
29 of 34 participants with sequence length of about 7000 frames. Some
examples from this database is shown in Figure \ref{fig:database}.

\section{Results}

In this section we report the results based on two expression datasets.
Figure \ref{fig:result1} shows the number of iterations needs for
convergence for each dataset. As can be seen in compare to ordinary
{GCRF} our proposed model converges very fast even for a long time
dataset (RU-FACS). Table \ref{table:result1} compares the area under
ROC curve for the proposed databases. As can be seen the proposed
model outperforms ordinary GCRF on two datasets.

\section{Conclusion}

In this paper the problem of temporal analysis is addressed. Many
methods have been proposed in literature that have their own advantages
and disadvantages. one of these methods which has been well studied
in recognition and detection is GCRF. The problem with this method
is the high computational cost for learning parameters. In this work
we tackle this problem and we introduce a new model based on using
ADMM. We evaluated our works on two publicly datasets and we show
how our method converge fast in compare to ordinary GCRF. Also our
proposed model outperforms GCRF.

\begin{table}
\begin{centering}
\label{result1-1} \includegraphics[scale=0.8]{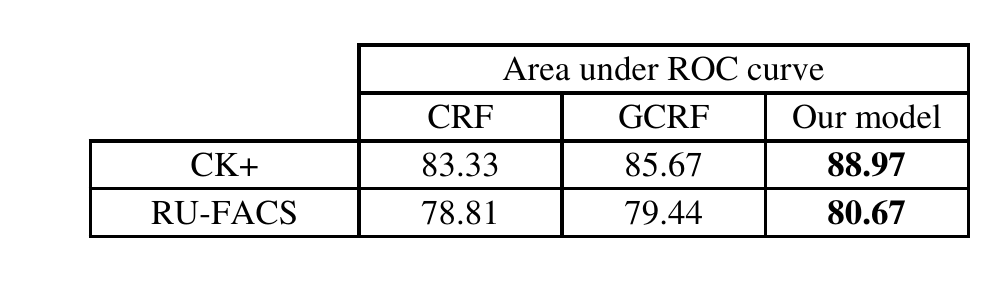} 
\par\end{centering}

\caption{Area under ROC curve for two expression databases}

\label{table:result1}
\end{table}

{\small{}\bibliographystyle{ieee}
\bibliography{refrence}
 }{\small \par}
\end{document}